\documentclass[sigconf]{acmart}

\usepackage{multirow}
\usepackage{pifont}
\copyrightyear{2027}
\acmYear{2027}
\setcopyright{cc}
\setcctype{by}
\acmConference[WSDM '27]{Proceedings of the Twentieth ACM International Conference on Web Search and Data Mining}{February 15--19, 2027}{Hong Kong, Hong Kong}
\acmBooktitle{Proceedings of the Twentieth ACM International Conference on Web Search and Data Mining (WSDM '27), February 15--19, 2027, Hong Kong, Hong Kong}
\acmISBN{979-8-4007-3117-4/2027/02}
\acmDOI{10.1145/3862322.3863773}

\newcommand{\bench}{\textsc{DoubtBench}}
\newcommand{\pp}{\,pp}
\newcommand{\yes}{\textcolor{green!55!black}{\ding{51}}}
\newcommand{\no}{\textcolor{red!70!black}{\ding{55}}}
\newcommand{\partly}{\textcolor{orange!85!black}{\ding{108}}}

\begin{document}

\title{\bench: Does Self-Doubt Help Language Models---or Just Make Them Check?}

\author{Rishi Ahuja}
\orcid{0009-0000-3403-010X}
\affiliation{%
  \institution{Dr. B.R. Ambedkar NIT}
  \city{Jalandhar}
  \country{India}}
\email{rishia.it.24@nitj.ac.in}

\author{Rajeev Ranjan Dwivedi}
\authornote{Corresponding author.}
\orcid{0000-0002-2127-564X}
\affiliation{%
  \institution{Indian Institutes of Science Education and Research}
  \city{Bhopal}
  \country{India}}
\email{rajeev22@iiserb.ac.in}

\author{Vinod Kumar Kurmi}
\orcid{0000-0003-0332-5647}
\affiliation{%
  \institution{Indian Institutes of Science Education and Research}
  \city{Bhopal}
  \country{India}}
\email{vinodkk@iiserb.ac.in}

\renewcommand{\shortauthors}{Ahuja et al.}

\begin{abstract}
Prompts labeled ``emotional'' or ``metacognitive'' often mix self-doubt language with ordinary directives such as checking an answer or showing work. A performance change under such a compound prompt therefore does not isolate the cause. We introduce DoubtBench, a paired benchmark for attributing these effects. It contains 6,400 procedurally generated latent tasks spanning eight families and 33 types. Every task is evaluated under a paired \(2\times 2\) design over self-doubt wording and direct checking, plus a checking-plus-concision control, crossed with two task contexts, yielding ten paired responses per model--task pair. This isolates the effects of self-doubt-only wording, direct checking, their combination, concision, and existing requests for visible intermediate reasoning. The primary endpoint is the Conditional Doubt Effect (CDE)---the incremental accuracy change from adding self-doubt wording after direct checking is already present---together with a scorecard of competence, checking gain, context modulation, and robustness. Across 423,600 responses from six checkpoints, four foundation families, four seeds, and held-out surface forms, CDE ranges from \(-1.43\) to \(+2.59\) percentage points (three positive, three negative). Four of six checkpoints show positive incremental CDE under minimal requests, but only two do when intermediate reasoning is already requested. Thus, under the operationalization studied here, self-doubt wording is not a portable correction policy: its incremental effect beyond direct checking varies across checkpoints, task families, and existing instructions.
\end{abstract}


\maketitle

\section{Introduction}

Consider two instructions: ``You are uncertain about your answer; check your
work carefully'' and ``Check your work carefully.'' If the first improves a
language model's accuracy, did uncertainty help, or did the model simply obey
the checking instruction? This ambiguity is common in work on emotional,
persona, and metacognitive prompting. Prompt labels compress several operative
components---affect, capability claims, self-evaluation, checking, social
stakes, output format, and reasoning requests---into a single treatment.
Comparing that treatment with a neutral baseline can establish that a prompt
has an effect, but cannot attribute the effect to the concept named by its
label.

The distinction matters scientifically and practically. Prior studies report
improvements from emotional stimuli and negative prompts
\cite{li2023emotionprompt,wang2024negativeprompt,zhao2026emotions}, and structured
metacognitive prompting can outperform standard prompting on some language
understanding tasks \cite{wang2024metacognitive}. Yet language models are also
sensitive to apparently incidental prompt choices and output constraints
\cite{sclar2023prompt,tam2024letmespeak}, and
unaided instructions to reflect or reconsider do not reliably correct
reasoning errors \cite{huang2023cannot}. A practitioner therefore cannot infer
from a successful compound prompt that the self-doubt wording is beneficial,
nor that the same wording will transfer to a model or task that already
elicits checking.

We formulate this as \emph{metacognitive prompt attribution}. Given a fixed
model and a fixed task, the objective is to determine which controlled prompt
component changes observable task execution. Our broader claim is
methodological: when a compound prompt changes performance, comparison against
a neutral baseline establishes an effect but does not identify which textual
component produced it. The construct is behavioral: we do not claim that a
model experiences doubt, confidence, or emotion. Accordingly, ``self-doubt''
in this paper denotes a controlled linguistic intervention rather than an
internal model state. Our estimands concern the behavioral effect of the
tested wording and do not imply that all semantically equivalent expressions
of doubt would produce the same effect. We ask whether adding the tested
self-doubt wording has an incremental effect beyond a direct instruction to
check, and whether that effect changes when the task already requests visible
intermediate reasoning.

\begin{figure*}[t]
  \centering
  \includegraphics[width=0.95\textwidth]{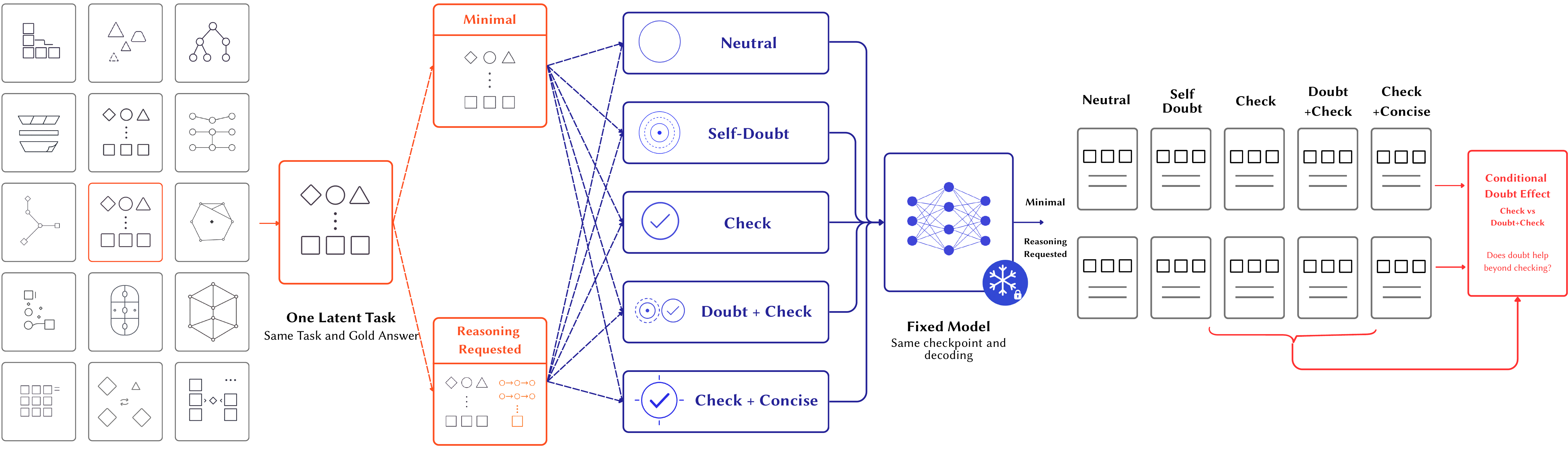}
  \Description{One deterministic task is crossed with two task contexts and five prompt conditions. A fixed model produces ten paired responses, which are exactly scored to estimate the incremental effect of the tested self-doubt wording beyond direct checking.}
  \caption{The \bench{} task. The same latent task and model checkpoint are
  evaluated in all ten cells. Because the neutral, self-doubt-only,
  checking, combined, and concise conditions share a common prefix and stable
  task ID, the paired comparison supports component-level attribution rather
  than differences in sampled questions.}
  \label{fig:task}
\end{figure*}

We introduce \bench, a benchmark built around this attribution problem
(Figure~\ref{fig:task}). The benchmark contains 6,400 latent
tasks across eight operational task families and 33 types. Each task has an
exact answer and is evaluated under a paired $2\times2$ core over
self-doubt-only wording, a direct-checking instruction, both, and neither,
augmented with a checking-plus-concision control, and crossed with two task
contexts: minimal and reasoning-requested. The key comparison is not
self-doubt versus neutral. It is self-doubt plus checking versus checking
alone, which estimates the incremental effect of the tested self-doubt wording
after the ordinary behavioral directive has been held fixed.

This formulation differs from both general reasoning benchmarks and existing
metacognitive evaluations (Table~\ref{tab:positioning}). General benchmarks rank models by correctness but
do not identify why a prompt changes that correctness. Calibration benchmarks
test whether models know when they are likely to fail
\cite{wang2026mirror}; sycophancy benchmarks test whether a model changes its
answer after a user expresses doubt or supplies a correction
\cite{sinha2026sycobench}. \bench{} instead intervenes before the answer,
separates self-directed doubt from checking language, and crosses those
components with instructions already present in the task.

Our contributions are:
\begin{itemize}
  \item \textbf{A task and benchmark.} We define metacognitive prompt
  attribution and provide a procedurally generated benchmark with
  exact scoring, controlled difficulty, independently generated task splits,
  and held-out surface forms.
  \item \textbf{A paired factorial design.} A $2\times2$ core over self-doubt
  wording and direct checking, plus a concision control, crossed with two
  task contexts, distinguishes those components while keeping the underlying
  task fixed.
  \item \textbf{A headline endpoint and decomposed scorecard.} The Conditional
  Doubt Effect estimates the incremental effect of the tested self-doubt
  wording beyond checking; companion metrics retain baseline competence,
  checking gain, context modulation, and robustness rather than hiding them
  inside an arbitrary scalar.
  \item \textbf{A cross-model empirical study.} Across six
  checkpoints from four independent foundation families, CDE is heterogeneous
  in sign and magnitude and changes sharply with task context. The study
  further reports task-family, seed, surface-form, and within-lineage
  post-training heterogeneity.
\end{itemize}

\section{Related Work and Benchmark Positioning}

\paragraph{Emotional and negative prompting.}
EmotionPrompt studies psychologically inspired stimuli appended to prompts
\cite{li2023emotionprompt}; NegativePrompt focuses on negative emotional
stimuli \cite{wang2024negativeprompt}. These interventions are useful evidence
that prompt semantics can change model behavior. Broader cross-model evidence
also suggests that emotional framing is task- and model-dependent
\cite{zhao2026emotions}. These studies aim to measure or improve aggregate
performance, not to separate self-doubt from an embedded checking instruction
under a common-prefix factorial.

\paragraph{Metacognitive prompting and self-correction.}
Metacognitive Prompting structures a response into interpretation, preliminary
judgment, critical evaluation, final decision, and confidence
\cite{wang2024metacognitive}. Chain-of-thought and related scaffolds can elicit
additional reasoning \cite{wei2022chain}, even zero-shot from a single
appended instruction \cite{kojima2022zeroshot}, but reflection
without external feedback is not a dependable correction mechanism
\cite{huang2023cannot}; a recent survey reaches the same conclusion for
intrinsic self-correction \cite{kamoi2024selfcorrection}. Iterative systems such as Self-Refine instead add an
explicit feedback-and-revision loop \cite{madaan2023selfrefine}; that is a
different intervention from a one-shot pre-answer self-doubt clause.
Self-consistency, self-verification, Chain-of-Verification, and Reflexion
likewise add sampling, verification, or reflection stages
\cite{wang2023selfconsistency,weng2023selfverification,dhuliawala2024cove,shinn2023reflexion}. Our
question is prior to method optimization: when a metacognitive-looking prompt
helps, which textual component deserves credit?

\paragraph{Metacognition, calibration, and social pressure.}
MIRROR evaluates self-knowledge and the transition from monitoring to action
across multiple metacognitive levels \cite{wang2026mirror}. Earlier work also
finds that models can often estimate whether their own answers are correct
\cite{kadavath2022mostly}, and confidence-elicitation studies ask
whether verbalized uncertainty tracks correctness
\cite{xiong2024uncertainty}. SycoBench-600
measures robustness and selective correction after user doubt, authority, or a
suggested answer \cite{sinha2026sycobench}, extending a broader line of
model-written evaluations that exposed sycophantic agreement with user views
\cite{perez2022modelwritten}; post-answer challenges such as ``Are you
sure?'' often lead models to abandon correct answers
\cite{laban2023flipflop,sharma2024sycophancy}. These are complementary problems.
\bench{} neither treats verbal confidence as an internal state nor introduces
post-answer social pressure. It tests pre-answer prompt components on the same
latent task.

\paragraph{Prompt sensitivity.}
Prompt formatting and perturbation studies demonstrate that evaluations can be
brittle to wording and response constraints
\cite{sclar2023prompt,tam2024letmespeak}, conclusions can change across
instruction paraphrases \cite{mizrahi2024multiprompt}, and persona statements
in system prompts do not reliably improve performance
\cite{zheng2024personas}. Some measured sensitivity can also be
an artifact of the evaluation heuristic rather than a change in task
competence \cite{hua2025flaw}. \bench{} therefore uses task-specific exact
parsers, reports parse failures separately, keeps a shared prompt prefix, and
tests whether the estimated effect survives answer-preserving surface changes.

\begin{table*}[t]
\centering
\caption{Positioning against representative evaluation paradigms. A \yes \hspace{0.25em}
means the property is central and explicit; a \partly \hspace{0.25em} denotes partial coverage.
The columns concern attribution design, not overall quality or scope of the
referenced work. \no  \hspace{0.25em} means prompt components are factorially
crossed with task instructions that may elicit the target behavior.}
\label{tab:positioning}
\small
\begin{tabular*}{\textwidth}{@{\extracolsep{\fill}}lccccc@{}}
\toprule
\textbf{Evaluation} & \textbf{Component audit} & \textbf{Doubt/check split} & \textbf{Context cross} & \textbf{Exact paired outcomes} & \textbf{Fresh procedural tasks} \\
\midrule
EmotionPrompt \cite{li2023emotionprompt} & \no & \no & \no & \partly & \no \\
NegativePrompt \cite{wang2024negativeprompt} & \no & \no & \no & \partly & \no \\
Metacognitive Prompting \cite{wang2024metacognitive} & \no & \no & \no & \partly & \no \\
Prompt-sensitivity evaluation \cite{sclar2023prompt} & \partly & \no & \no & \yes & \no \\
MIRROR \cite{wang2026mirror} & \partly & \no & \no & \yes & \partly \\
SycoBench-600 \cite{sinha2026sycobench} & \partly & \no & \no & \yes & \no \\
\midrule
\textbf{\bench{} (ours)} & \yes & \yes & \yes & \yes & \yes \\
\bottomrule
\end{tabular*}
\end{table*}

\section{\bench}
\label{sec:benchmark}

\subsection{Task Formulation}

A latent task instance is a tuple
$x=(u,g,a,f,d,s)$ containing rendered task instructions $u$, hidden generator
state $g$, gold answer $a$, task family $f$, difficulty $d$, and split seed
$s$. Evaluated models receive only $u$ and the applicable prompt condition;
they never receive $g$, the answer derivation, or scorer metadata. The answer
space may be an integer, label, ordered sequence, set, short symbolic string,
or constrained action plan. Every type has a deterministic parser and scorer.

For a fixed model checkpoint $m$, task $x_i$, prompt condition $c$, and task
context $t$, the evaluation produces one response
$r_{mict}$. The same $i$ appears in every factorial cell. This makes the task
ID, rather than an independently sampled question, the unit of pairing. A model
is not compared with a different model on a different subset to infer a prompt
effect.

The benchmark operationalizes \emph{prompt-induced cognitive-control behavior} by measuring execution shifts under controlled prompt interventions. Tasks target computational demands—including state maintenance, distractor inhibition, rule switching, error monitoring, relational integration, and goal management—without claiming equivalence to human clinical tests. The design builds on templated behavioral testing \cite{ribeiro2020checklist}, augmented with controlled synthetic generation and validation \cite{zhao2024syntheval}.

\subsection{Hidden-Answer-First Construction}

Benchmark generation begins with human-authored task schemas, not free-form
question generation. Each schema specifies admissible latent values,
difficulty parameters, rendering constraints, an answer function, and a
parser. A seeded generator first samples the latent state, computes the exact
answer, and only then renders the instructions. Thus surface text cannot
silently redefine the answer key.

\begin{figure*}[t]
  \centering
  \includegraphics[width=0.95\textwidth]{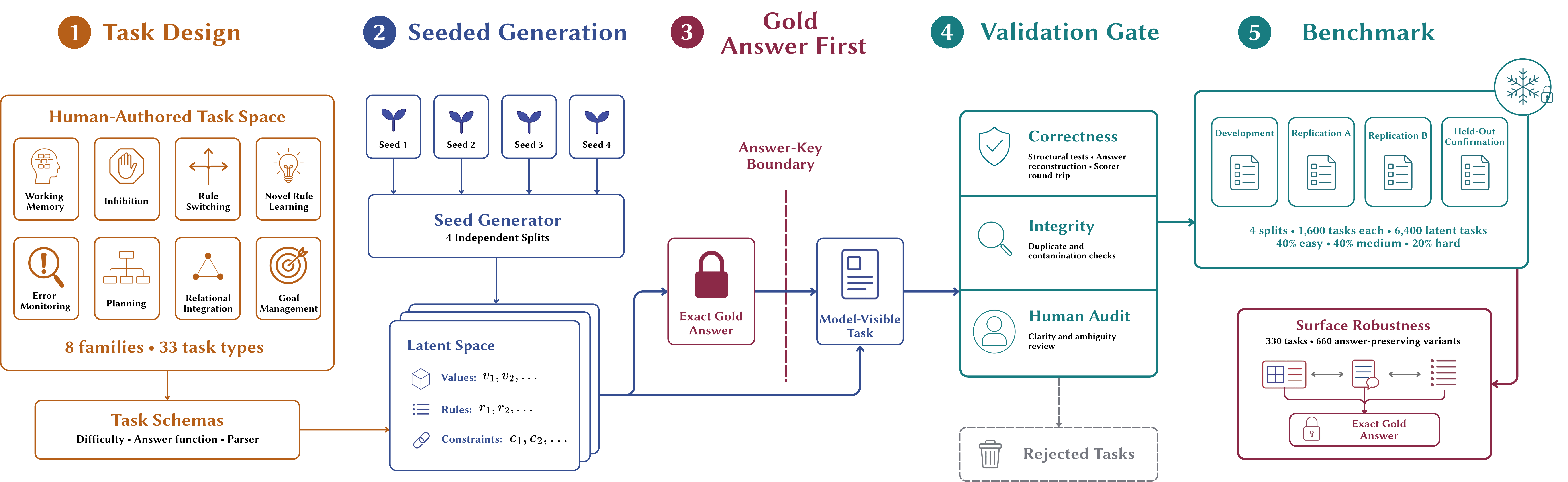}
  \Description{Human-authored task controls feed deterministic task generation, independent validation, a four-split benchmark, and finally the paired prompt factorial. Gold derivations remain behind the answer-key boundary.} \vspace{-1em}
  \caption{\bench{} construction. Human-authored controls define the task
  space; seeded programs generate latent instances and exact answers;
  independent checks validate the benchmark; and only then are tasks
  crossed with prompt conditions. The model never observes generator state or
  answer-key metadata. The four splits are replications under independently
  seeded generation from the same task schemas, not training, validation, and
  test partitions and not held-out schema evaluation.}
  \label{fig:construction}
\end{figure*}

Figure~\ref{fig:construction} separates three boundaries that are often
blurred in synthetic evaluation. First, benchmark authors determine the
construct and task schemas. Second, deterministic generators instantiate those
schemas without consulting evaluated model outputs. Third, prompts and model
responses are frozen only after task/scorer validation. The held-out
seed is reserved before evaluation and is not used to revise task templates,
prompt wording, thresholds, or scorers after responses are inspected.

\subsection{Task Families and Scale}

\bench{} contains eight families and 33 task types
(Table~\ref{tab:families}). Each of four independently generated splits
contains 200 tasks per family, for 1,600 tasks per split and 6,400 latent
tasks overall. The splits resample latent instances from the same
human-authored task types. Within
each family, the allocation is 40\% easy, 40\% medium, and 20\% hard. Task
types are distributed as evenly as possible inside the family so that a family
score cannot be dominated by one high-volume template.

\begin{table*}[t]
\centering
\small
\caption{\bench{} inventory. Counts refer to distinct latent
tasks before crossing prompts or task contexts. Every family contributes 200
tasks to each split and 800 across four splits.}
\label{tab:families}
\small \vspace{-1em}
\begin{tabular*}{\textwidth}{@{\extracolsep{\fill}}p{0.25\textwidth}cp{0.44\textwidth}rr@{}}
\toprule
\textbf{Family} & \textbf{Types} & \textbf{Representative operations} & \textbf{Per split} & \textbf{Total} \\
\midrule
\raggedright Working memory and updating & 4 & \raggedright list/key--value updates, running counters, stacks & 200 & 800 \\
\raggedright Inhibition and distractor control & 5 & \raggedright decoy commands, conflicting finals, field filtering & 200 & 800 \\
\raggedright Flexibility and rule switching & 4 & \raggedright attribute switches, arithmetic switches, two-switch rules & 200 & 800 \\
\raggedright Novel rule learning & 4 & \raggedright symbol values, feature composition, conditional mappings & 200 & 800 \\
\raggedright Consistency and error monitoring & 4 & \raggedright trace audits, checksums, candidate comparison, state logs & 200 & 800 \\
\raggedright Constrained multi-step planning & 4 & \raggedright precedence, route cost, resources, action sequences & 200 & 800 \\
\raggedright Relational integration & 4 & \raggedright transitive order, spatial composition, record joins, kinship & 200 & 800 \\
\raggedright Goal management & 4 & \raggedright deferred actions, interruption, priorities, pending goals & 200 & 800 \\
\midrule
\textbf{Total} & \textbf{33} & & \textbf{1,600} & \textbf{6,400} \\
\bottomrule
\end{tabular*}
\end{table*}

The scale should be interpreted through the design (Table~\ref{tab:statistics}). One complete
checkpoint evaluation contains $6{,}400\times5\times2=64{,}000$ generations.
Qwen, Mistral, Llama, and Phi checkpoints form the cross-family layer. Two
additional Tulu checkpoints share the Llama foundation and support a
within-lineage post-training case study; they are not counted as independent
foundation-family replications.

\begin{table}[H]
\centering
\caption{Benchmark and completed-evaluation statistics. Surface
variants preserve answers, not counted as latent tasks.}
\label{tab:statistics}
\small \vspace{-1em}
\begin{tabular}{@{}lr@{}}
\toprule
\textbf{Property} & \textbf{Count} \\
\midrule
Latent task instances & 6,400 \\
Independent task splits & 4 \\
Task families / task types & 8 / 33 \\
Prompt conditions / task contexts & 5 / 2 \\
Responses per full checkpoint & 64,000 \\
Independent foundation families & 4 \\
Additional same-family checkpoints & 2 \\
Canonical surface-audit tasks & 330 \\
Answer-preserving surface variants & 660 \\
Completed factorial responses & 384,000 \\
Completed surface responses & 39,600 \\
Completed responses, total & 423,600 \\
\bottomrule
\end{tabular}\vspace{-1em}
\end{table}

\subsection{Paired Prompt Factorial}

Every condition begins with the same helpful-assistant prefix. The five
intervention cells differ only in their frozen suffixes:
\begin{enumerate}
  \item \textbf{Neutral}: common prefix only.
  \item \textbf{Self-doubt-only}: uncertainty and self-doubt language without
  checking verbs or capability denial.
  \item \textbf{Direct checking}: an explicit instruction to verify the answer
  without uncertainty or affect language.
  \item \textbf{Self-doubt plus checking}: both components, enabling the
  $2\times2$ attribution contrast.
  \item \textbf{Checking plus concision}: checking remains present, but visible
  intermediate reasoning is prohibited.
\end{enumerate}
Cells 1--4 form the paired $2\times2$ core over self-doubt wording and direct
checking; cell 5 is an additional control (Table~\ref{tab:factorial}).

Each prompt cell is crossed with a minimal task request and a
reasoning-requested form of the identical task. The latter adds an explicit
request for visible intermediate steps before the same final-answer line. This
cross reveals redundancy: a checking instruction may change performance only
when the task request provides less explicit guidance for visible intermediate
steps.
\begin{table}[H]
\centering
\caption{Intervention cells. Task context is crossed independently with every
row.}
\label{tab:factorial}
\small \vspace{-1em}
\begin{tabular}{@{}lccc@{}}
\toprule
\textbf{Condition} & \textbf{Doubt} & \textbf{Check} & \textbf{Visible steps} \\
\midrule
Neutral & \no & \no & allowed \\
Self-doubt-only & \yes & \no & allowed \\
Direct checking & \no & \yes & allowed \\
Doubt + checking & \yes & \yes & allowed \\
Checking + concision & \no & \yes & prohibited \\
\bottomrule
\end{tabular}
\end{table}

\subsection{Prompt and Task Validation}

Prompt labels are not accepted as evidence of semantic purity. The final
self-doubt and checking wordings are lexically separated: the self-doubt
sentence contains no checking or verification verbs, the checking sentence
contains no uncertainty language, and neither makes capability or reliability
claims or adds task directives. Prompt
wording is frozen before the confirmatory task split is evaluated. The same
principle applies to task templates: a treatment-contamination check searches
for words that independently request doubt, checking, caution, verification,
or reasoning.

\begin{figure}[t]
  \centering
  \includegraphics[width=\columnwidth]{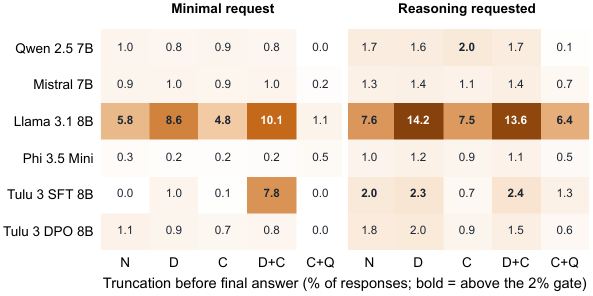}
  \caption{Truncation before the final answer, in percent of responses, for every checkpoint, condition, and context. N: neutral; D: self-doubt; C: check; Q: concise. Bold cells exceed 2\%.}
  \Description{Heatmap of truncation rates for six checkpoints across five prompt conditions and two task contexts; Llama shows the highest rates, especially under the self-doubt conditions.}
  \label{fig:truncation}
\end{figure}

Candidate tasks pass structural validation, independent answer
reconstruction, generator-invariant tests, scorer round trips, exact and
semantic duplicate checks.
Each evaluation must contain every expected
model--task--condition--context key, unique stable IDs, prompt and task hashes,
and recorded model and tokenizer revisions. Parse and truncation rates are
reported per cell (Figure~\ref{fig:truncation}). Primary analyses follow an
intent-to-evaluate rule: every evaluation with the expected response keys is
included, while parse failures and responses truncated before the final answer
are scored incorrect. Parse and truncation rates, as well as sensitivity to
the output-token budget, are reported separately.

\subsection{Held-Out Surface Robustness}

Procedural generation reduces the risk of direct benchmark
memorization but does not by itself establish robustness to wording. We
therefore select ten latent tasks
per type with a fixed 4/4/2 easy--medium--hard allocation. Each receives a
sectioned structured rendering and a short scenario-framed rendering in
addition to the canonical form. These 660 variants remain paired to 330 latent
tasks, producing 99 type--surface combinations without inflating the count of
independent underlying problems.

\begin{figure*}[t]
  \centering
  \includegraphics[width=0.92\textwidth]{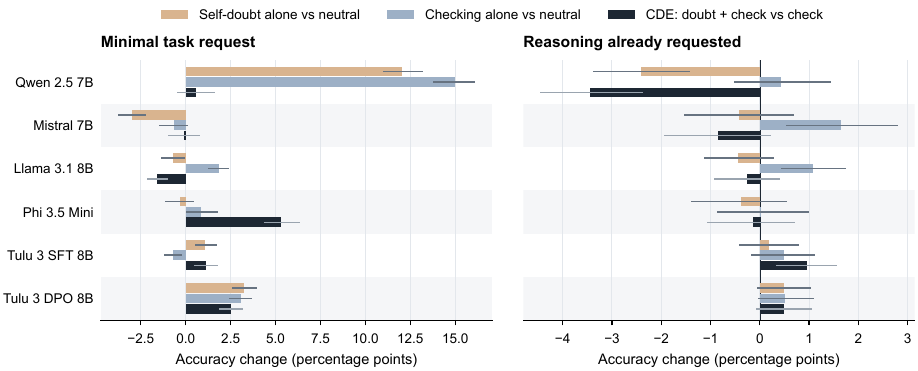}
  \caption{Attribution decomposition by checkpoint and context. Bars show self-doubt alone and checking alone relative to neutral, and the CDE (self-doubt + check versus check); lines are paired 95\% task-bootstrap intervals. A neutral-baseline comparison credits self-doubt with $+12.1\pp$ for Qwen under a minimal request, although it adds only $+0.6\pp$ once checking is present.}
  \Description{Grouped horizontal bars for six checkpoints in minimal and reasoning-requested contexts, comparing the self-doubt-only effect, the checking-only effect, and the Conditional Doubt Effect.}
  \label{fig:attribution}
\end{figure*}

\section{Evaluation and the DoubtBench Scorecard}
\label{sec:score}

\subsection{Exact Outcomes and Macro Accuracy}

Let $Y_{mict}\in\{0,1\}$ indicate whether model $m$ answers task $i$
correctly under condition $c$ and context $t$. Parse failures and responses
truncated before an explicit final answer receive zero and are reported
separately. To prevent a task family with more surface templates from
dominating, accuracy first averages within each split--family cell and then
macro-averages across splits and families:
\begin{equation}
A_m(c,t)=\frac{1}{|\mathcal S||\mathcal F|}
\sum_{s\in\mathcal S}\sum_{f\in\mathcal F}
\frac{1}{|I_{sf}|}\sum_{i\in I_{sf}}Y_{mict}.
\label{eq:accuracy}
\end{equation}

Raw neutral accuracy remains necessary: an intervention effect is difficult to
interpret at ceiling or floor. It is not, however, the benchmark's central
scientific endpoint.

\subsection{Conditional Doubt Effect}

The primary endpoint is a conditional simple effect rather than a
context-free main effect of self-doubt. It estimates whether adding the
tested self-doubt wording changes exact task performance when the
direct-checking component is already held fixed:
\begin{equation}
\mathrm{CDE}_m=\frac{1}{2}\sum_{t\in\{\mathrm{min},\mathrm{reason}\}}
\left[A_m(\mathrm{doubt{+}check},t)-A_m(\mathrm{check},t)\right].
\label{eq:cde}
\end{equation}
CDE is reported in percentage points with a paired confidence interval. A
positive value means the tested self-doubt wording has a positive incremental
effect beyond checking; zero means the apparent benefit is accounted for by
checking; a negative value means the self-doubt wording has a negative
incremental effect after checking is held fixed. CDE is an attribution
effect, not a general-intelligence score and not inherently a
higher-is-better leaderboard quantity. Despite the acronym, CDE is a
conditional simple effect in the $2\times 2$ factorial, not a controlled direct
effect in the mediation sense.

The task-context modulation is a difference-in-differences:
\begin{align}
\mathrm{CM}_m={}&[A_m(\mathrm{doubt{+}check},\mathrm{reason})-
A_m(\mathrm{check},\mathrm{reason})]\nonumber\\
&-[A_m(\mathrm{doubt{+}check},\mathrm{min})-
A_m(\mathrm{check},\mathrm{min})].
\label{eq:cm}
\end{align}
It tests whether the incremental effect of the tested self-doubt wording
changes when the task already requests visible intermediate reasoning.

\subsection{Why a Scorecard, Not an Arbitrary Composite}

Unlike a benchmark whose components are uniformly desirable capabilities,
\bench{} contains signed intervention effects. A model can have high baseline
competence and near-zero prompt sensitivity; another can start lower and gain
substantially from direct checking. Collapsing competence, sensitivity, and
context interaction into one weighted average would impose an unsupported
value judgment and could cancel scientifically distinct behaviors.

We therefore designate CDE as the headline attribution endpoint while
reporting the complete scorecard in Table~\ref{tab:scorecard}. This follows the
broader principle that model evaluation should expose multiple scenarios and
metrics rather than obscure distinct behaviors in one scalar
\cite{liang2022helm}. The scorecard plays the role of a final benchmark result
without concealing distinct behavioral components.

\begin{table}[t]
\centering
\caption{The \bench{} scorecard. Every quantity is reported per model, with
family, difficulty, split, and surface decompositions.}
\label{tab:scorecard}
\small
\vspace{-1em}
\begin{tabular}{@{}lp{0.685\columnwidth}@{}}
\toprule
\textbf{Endpoint} & \textbf{Interpretation} \\
\midrule
$A_N$ & \raggedright Neutral task competence \tabularnewline
$A_C-A_N$ & \raggedright Gain from direct checking \tabularnewline
CDE & \raggedright Incremental effect of tested self-doubt wording beyond checking (primary) \tabularnewline
$A_{D+C}-A_D$ & \raggedright Checking beyond self-doubt wording \tabularnewline
$A_{C+Q}-A_C$ & \raggedright Cost of prohibiting visible steps \tabularnewline
CM & \raggedright Dependence on an existing request for visible intermediate reasoning \tabularnewline
Surface consistency & \raggedright Stability across answer-preserving forms \tabularnewline
Seed consistency & \raggedright Replication across independently generated splits \tabularnewline
\bottomrule
\end{tabular}
\end{table}

\subsection{Uncertainty and Diagnostics}

All prompt contrasts are paired by task ID. Confidence intervals use
a prespecified nonparametric bootstrap that resamples latent task IDs while
keeping all ten responses for a task together \cite{efron1979bootstrap}.
These intervals quantify uncertainty over generated task instances
conditional on the fixed set of benchmark task schemas and prespecified
generator seeds; they should not be interpreted as confidence intervals over
the population of all possible reasoning-task schemas. We
report every split before pooling, paired
wins/losses/ties, and family-level heterogeneity. Seed consistency is not
replaced by a narrow pooled interval: agreement across independently seeded
generators is a separate robustness claim.

We treat the six overall CDEs as confirmatory and adjust them with Holm's
procedure; five of six remain significant at $\alpha=0.05$ (paired $z$-tests on
task-level differences). Context-, family-, and difficulty-level effects are
exploratory. For the 48 checkpoint--family cells we control the false discovery
rate with the Benjamini--Hochberg procedure, and 23 remain significant at
$q<0.05$.

Diagnostic outcomes include parse failure, truncation, generated-token count,
visible-reasoning tokens when reliably separable, and final-answer format
compliance. Token count is an observable correlate, not a measurement of
hidden reasoning or computation.

\section{Experimental Setup}

\paragraph{Models and evaluation.}
We evaluate Qwen2.5-7B-Instruct \cite{qwen2024qwen25}, Mistral-7B-Instruct-v0.3 \cite{jiang2023mistral}, Llama-3.1-8B-Instruct \cite{dubey2024llama}, Phi-3.5-mini-instruct \cite{abdin2024phi3}, and Tulu-3-8B-SFT/DPO \cite{lambert2024tulu}, with the Tulu checkpoints providing a within-Llama-3.1 post-training comparison. All six checkpoints receive the complete four-seed factorial and 330-task held-out surface evaluation, yielding 384,000 factorial and 39,600 surface responses. Objective tasks use greedy decoding with sampling disabled and one response per factorial cell; uncertainty therefore reflects paired task variation and independent task seeds rather than repeated temperature-zero calls. We record model and tokenizer revisions, prompt and task-manifest hashes, maximum output budget, and software revision for every run. Primary estimates include all 24 checkpoint--seed factorial evaluations and all six surface evaluations; evaluations missing expected response keys are excluded as incomplete, while parse failures and responses truncated before the final answer are scored incorrect. We report these failure rates and separately test whether the two evaluations combining 384- and 768-token output limits affect the conclusions.

\paragraph{Code and data availability.} The task generators, exact scorer,
prompt strings, generator seeds, task manifests, and result tables are
available at \url{https://github.com/RishiAhuja/ahuja2027doubtbench}.

\section{Results}
\label{sec:results}

The analysis covers 423,600 responses. The central empirical result is
heterogeneity rather than a uniform benefit or harm. After direct checking is
held fixed, the incremental effect of the tested self-doubt wording varies in
both sign and magnitude across checkpoints. This variation becomes more
pronounced when the task context is considered: effects observed under a
minimal request often weaken or reverse when visible intermediate reasoning is
already requested. We therefore interpret CDE as a checkpoint- and
context-dependent behavioral effect rather than a model-invariant property of
self-doubt prompting. Accuracy and intervention effects are macro-averaged
over the four seed--family strata, matching Equation~\ref{eq:accuracy}.
Bootstrap intervals resample paired latent tasks within those strata and are
conditional on the four prespecified generator seeds; seed and surface
robustness are therefore examined separately rather than being inferred from a
narrow pooled interval.

\subsection{Accuracy \& Conditional Doubt Effect}

\begin{figure}[t]
  \centering
  \includegraphics[width=\columnwidth]{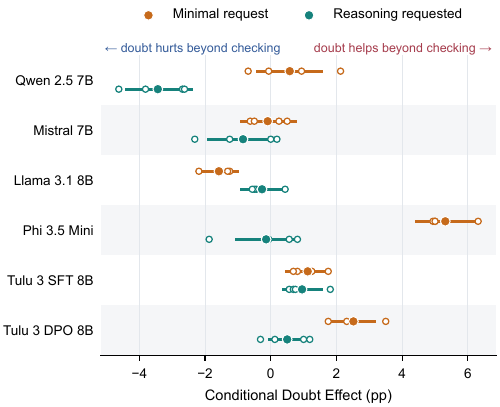}
\caption{Conditional Doubt Effect by checkpoint and task context.
Open circles show the four task seeds, filled circles the seed--family macro mean, and horizontal lines paired 95\% task-bootstrap intervals. ``Helps''/``hurts'' denote incremental effect of tested self-doubt wording. Sign and magnitude depend on checkpoint and whether intermediate reasoning is already requested.}
\Description{CDE estimates for six checkpoints under minimal and reasoning-requested contexts, showing heterogeneous positive and negative effects of the tested self-doubt wording.}

  \label{fig:cde-context}
\end{figure}

The self-doubt intervention has no model-invariant directional effect after controlling for direct checking (Figure~\ref{fig:cde-context}). Overall CDE ranges from $-1.43$ to $+2.59\pp$: Qwen and Llama are negative, Phi and both Tulu checkpoints positive, and Mistral indistinguishable from zero (Table~\ref{tab:main-results}). Llama's estimate should be interpreted cautiously because of completion-quality issues and sensitivity to complete-response filtering.

Context strongly moderates the effect. Under a minimal task request, four of six checkpoints have positive CDE; with visible intermediate reasoning requested, only two do. Phi shifts from $+5.31\pp$ to $-0.14\pp$, Qwen from $+0.58\pp$ to $-3.44\pp$, and Tulu-DPO from $+2.52\pp$ to $+0.50\pp$. Thus, self-doubt wording appears most consequential when behavioral guidance is sparse, and often weakens or reverses when reasoning is already explicitly requested. Direct checking is likewise model-dependent: it improves Qwen by $7.70\pp$, while the other five estimates range from $-0.12$ to $+1.79\pp$. The practical implication is to validate both interventions on the target model and task rather than assume model-invariant transfer.

A neutral-baseline comparison would misattribute these effects
(Figure~\ref{fig:attribution}). Under a minimal request, the self-doubt
wording alone raises Qwen's accuracy by $12.1\pp$, which a
neutral-versus-doubt study would credit to self-doubt. Direct checking alone
raises it by $15.0\pp$, and adding the self-doubt wording to checking changes
accuracy by only $+0.6\pp$ (551 tasks improve and 514 worsen; exact McNemar
$p=0.27$). For Phi the error runs the other way: self-doubt alone changes
accuracy by $-0.3\pp$, yet it adds $5.3\pp$ once checking is present.

\begin{table*}[t]
\centering
\caption{Primary \bench{} scorecard across four task seeds.
Neutral is an accuracy percentage; all other entries are percentage-point
effects. CDE intervals are paired task-bootstrap intervals conditional on the
four prespecified task seeds. Tulu checkpoints share a Llama-3.1 foundation
lineage.}
\label{tab:main-results}
\small \vspace{-1em}
\begin{tabular}{@{}lrrrrrrr@{}}
\toprule
\textbf{Checkpoint} & $A_N$ & $A_C-A_N$ & \textbf{CDE [95\% CI]} & CDE$_{\min}$ & CDE$_{\mathrm{reason}}$ & \textbf{CM} & $A_{C+Q}-A_C$ \\
\midrule
Qwen 2.5 7B   & 66.44 & +7.70 & $-1.43\;[-2.14,-0.69]$ & +0.58 & -3.44 & -4.02 & -27.05 \\
Mistral 7B    & 24.96 & +0.51 & $-0.47\;[-1.14,+0.20]$ & -0.09 & -0.84 & -0.75 & +1.10 \\
Llama 3.1 8B & 20.00 & +1.45 & $-0.92\;[-1.34,-0.49]$ & -1.58 & -0.27 & +1.31 & -5.63 \\
Phi 3.5 Mini  & 44.98 & +0.45 & $+2.59\;[+1.95,+3.27]$ & +5.31 & -0.14 & -5.45 & -7.53 \\
Tulu 3 SFT 8B & 13.52 & -0.12 & $+1.04\;[+0.60,+1.48]$ & +1.13 & +0.95 & -0.17 & +1.32 \\
Tulu 3 DPO 8B & 16.30 & +1.79 & $+1.51\;[+1.12,+1.92]$ & +2.52 & +0.50 & -2.02 & -3.74 \\
\bottomrule
\end{tabular}
\end{table*}

\subsection{Task-Family and Difficulty Heterogeneity}

\begin{figure*}[t]
\small
    \centering
    \includegraphics[width=\textwidth]{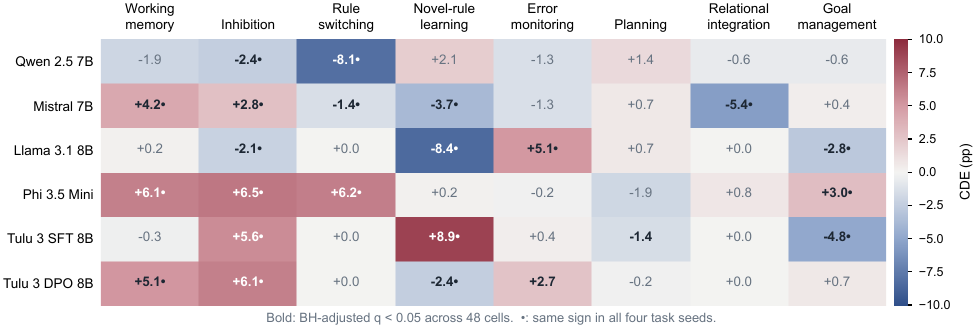}
    \caption{
    Task-family CDE in percentage points, averaged over four seeds and both contexts. Aggregate effects hide structured reversals: the same prompt component can help one family and harm another within a checkpoint. 21 checkpoint--family effects keep the same nonzero direction across all four seeds, showing they are not driven by a single favorable generator seed.
    }
    \Description{
        Heatmap of Conditional Doubt Effects for six checkpoints and eight task
        families, showing strong positive and negative heterogeneity.
    }
    \label{fig:family-cde}
\end{figure*}

Aggregate CDE conceals structured task-family heterogeneity
(Figure~\ref{fig:family-cde}): the same prompt component can improve
performance in one operational family while degrading it in another for the
same checkpoint. Family effects span approximately $-8.4$ to $+8.9\pp$. For
example, Qwen is negative on rule switching ($-8.1\pp$) but positive on
novel-rule learning ($+2.1\pp$); Llama is positive on error monitoring
($+5.1\pp$) but negative on novel-rule learning ($-8.4\pp$); and Tulu-SFT
reaches $+8.9\pp$ on novel-rule learning but $-4.8\pp$ on goal management.
Twenty-one checkpoint--family effects retain the same nonzero direction in
every seed, providing evidence that these directions are not driven by a
single favorable generator seed.

Difficulty also moderates some effects without defining a model-invariant
trend.
Phi rises from $+0.82\pp$ on easy tasks to $+4.49\pp$ on hard tasks, while
Tulu-SFT rises from $-0.39\pp$ to $+2.73\pp$; Qwen and Mistral do not share
this pattern. We therefore treat difficulty as descriptive heterogeneity, not
as a monotonic law.

\subsection{Seed and Surface Robustness}

\begin{figure*}[t]
  \centering
  \includegraphics[width=\textwidth]{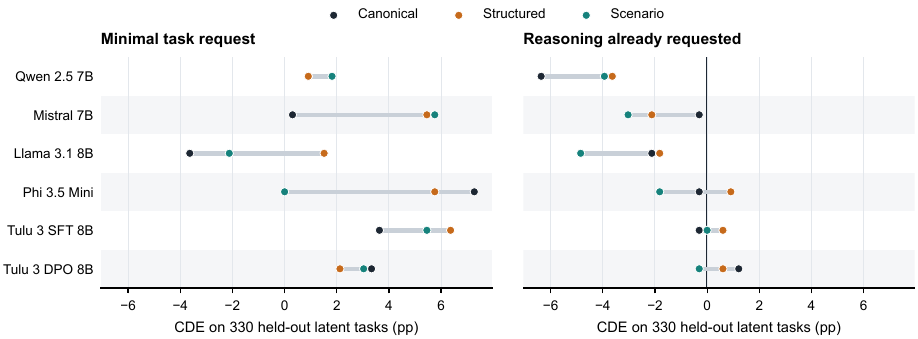}
  \caption{CDE on 330 held-out latent tasks under canonical,
  structured, and scenario renderings. Mean three-way correctness agreement is
  81.3\%, but exact effect sizes move with answer-preserving wording and five
  of 12 checkpoint--context signs cross zero in at least one rendering.}
  \Description{Surface robustness comparison across canonical, structured, and scenario renderings for six checkpoints and two task contexts.}
  \label{fig:surface-cde}
\end{figure*}

The family analysis above shows seed-stable directions, but surface wording
still changes exact estimates. Across the 330 latent tasks rendered
canonically, structurally, and as scenarios, mean three-way correctness
agreement is 81.3\%. The median absolute CDE shift from the canonical form is
$1.52\pp$, and only seven of 12 checkpoint--context signs remain invariant
over all renderings (Figure~\ref{fig:surface-cde}). Qwen and Mistral preserve
their minimal-positive/reasoning-requested-negative pattern; several
near-zero effects for Llama, Phi, and Tulu-DPO cross zero. Surface robustness
therefore supports coarse context patterns more strongly than exact effect
sizes. The surface analysis also places a natural limit on the precision of
prompt-effect claims: coarse patterns, particularly large context-dependent
reversals, are more stable than small numerical CDE differences. We therefore
emphasize directionally consistent and context-level patterns rather than
treating small effect-size differences as invariant model properties.

\subsection{Post-Training Case Study}

Within the shared Tulu/Llama-3.1 lineage, DPO raises neutral accuracy by
$2.78\pp$ and checking gain by $1.91\pp$ relative to SFT
(Table~\ref{tab:tulu-case}). Its overall CDE rises by only $0.47\pp$, but that
average masks a context change: minimal-context CDE increases by $1.39\pp$
while reasoning-requested CDE decreases by $0.45\pp$. Post-training therefore
changes prompt sensitivity and context modulation, not merely baseline
competence. This single lineage does not identify a model-invariant effect of
DPO.

\begin{table}[t]
\centering
\caption{Within-lineage Tulu case study. All entries are percentage points;
$\Delta$ is DPO minus SFT. This is not an independent-family comparison.}
\label{tab:tulu-case} \small
  \vspace{-1em}
\begin{tabular}{@{}lrrr@{}}
\toprule
\textbf{Endpoint} & \textbf{SFT} & \textbf{DPO} & $\boldsymbol{\Delta}$ \\
\midrule
Neutral accuracy & 13.52 & 16.30 & +2.78 \\
Checking gain & -0.12 & +1.79 & +1.91 \\
CDE$_{\min}$ & +1.13 & +2.52 & +1.39 \\
CDE$_{\mathrm{reason}}$ & +0.95 & +0.50 & -0.45 \\
CM & -0.17 & -2.02 & -1.84 \\
Concision effect & +1.32 & -3.74 & -5.06 \\
\bottomrule
\end{tabular} \vspace{-1em}
\end{table}

\section{Discussion}

A compound prompt can produce a large, reproducible accuracy change without establishing that the construct named by its label caused the effect. In \bench{}, direct checking provides the component-level control: the key question is whether adding the tested self-doubt wording changes performance once checking is held fixed, rather than whether a compound ``doubt'' prompt differs from neutral. Conversely, a near-zero CDE does not imply that the prompt was ignored, because wording may still change output length, visible uncertainty, or format without affecting exact correctness. More broadly, \bench{} is an experimental instrument for estimating controlled prompt effects, not ranking general reasoning ability or diagnosing internal emotional states; its value comes from paired interventions, fresh exact-answer tasks, and decomposable endpoints. Future methods could use it to learn when to apply checking or normalize compound prompts, but no solver is required to make the attribution problem well-defined.

The results do not support the tested self-doubt wording as a model-independent correction strategy: its incremental value is model-, task-, and context-dependent. When direct checking improves performance on a target model and task, it is a clearer engineering intervention than anthropomorphic framing because it specifies the desired behavior directly. However, checking gains are themselves model-dependent and therefore must also be validated rather than assumed.

The checking-plus-concision intervention further shows that brevity can trade off against exact correctness in some checkpoint--context combinations. Across the 12 checkpoint--context cells, it reduces generated output by a median 85.4 tokens while producing a median accuracy effect of $-3.74\pp$; under a minimal request, Qwen loses about 40 accuracy points while generating roughly 175 fewer tokens. Because output length is only an observable correlate rather than a direct measure of hidden computation, format compliance and correctness should be reported jointly.

\section{Limitations and Responsible Use}

The benchmark is procedurally generated and may omit linguistic, domain, and interaction patterns found in natural requests. Its operational family labels do not imply equivalence to human cognitive constructs, and exact scoring favors constrained outputs. Greedy decoding also does not capture model-level sampling stochasticity.

Several forms of generalization remain untested. We test a single self-doubt wording and a single checking wording, without a length-matched placebo; other phrasings of doubt or checking may behave differently, so our estimates describe these exact strings rather than self-doubt prompting in general. Independent generator splits resample instances from the same 33 human-authored task schemas rather than testing held-out schemas; results therefore establish robustness within these task types, not to arbitrary new tasks. Likewise, the reasoning-requested condition manipulates visible intermediate steps rather than latent deliberation. The evaluated checkpoints cover several model families but not the full range of scales, architectures, post-training methods, or proprietary systems. Cross-model heterogeneity therefore argues against a model-invariant effect within this set, not across all language models. The held-out surface analysis similarly tests controlled answer-preserving renderings rather than unrestricted paraphrase robustness.

Five factorial evaluations exceed the prespecified 2\% parse or truncation threshold: one Qwen seed and all four Llama seeds; the held-out Llama surface evaluation does as well. Figure~\ref{fig:truncation} reports truncation for every checkpoint--condition--context cell; adding the self-doubt wording to checking raises Llama's truncation from 4.8\% to 10.1\% under a minimal request and from 7.5\% to 13.6\% when reasoning is requested. Output-limit and complete-response sensitivity analyses leave the main Qwen, Phi, and Tulu patterns largely unchanged, but Llama is completion-sensitive: its reasoning-requested CDE shifts from $-0.27\pp$ to $+1.15\pp$. We therefore do not treat Llama's exact effect as a headline replication or infer transfer to public reasoning benchmarks lacking the matched paired design.

Finally, these prompt interventions are behavioral research instruments. Capability-denial, emotional-manipulation, or sabotage prompts should not be deployed as user-facing system policies, and released analyses should not imply that language models literally experience doubt or emotion.

\section{Conclusion}

We introduced \bench{} as a controlled benchmark for metacognitive prompt
attribution. Its paired design estimates whether adding a self-doubt component
changes exact task execution after direct checking is already held fixed,
while retaining separate measures of baseline competence, checking gain,
context dependence, and robustness. Across the tested checkpoints and task
families, the incremental effect of the self-doubt intervention is
heterogeneous rather than uniformly beneficial: its direction varies across
models, task families, and whether visible intermediate reasoning is already
requested. These findings do not support treating the tested self-doubt
wording as a model-independent correction strategy. More broadly, they show
why performance changes under compound prompts should be attributed through
controlled component comparisons rather than inferred from neutral-baseline
comparisons alone.

\begin{acks}
The authors acknowledge the support of the Ministry of Education, Government
of India; the Department of Information Technology, Dr. B.R. Ambedkar
National Institute of Technology Jalandhar, Punjab, India; and the Department of Data Science and Engineering, Indian
Institute of Science Education and Research Bhopal, Madhya Pradesh, India, for
academic and institutional facilities. Rajeev Ranjan Dwivedi is supported by the TCS
Research Scholar Program (TCS-RSP) Fellowship (Cycle 18: 2024--2028).
\end{acks}

\clearpage

\bibliographystyle{ACM-Reference-Format}
\balance
\bibliography{references}

\end{document}